\pdfoutput=1

\documentclass[11pt]{article}

\usepackage{EMNLP2023}

\usepackage{graphicx}
\usepackage{adjustbox}
\usepackage{times}
\usepackage{colortbl}
\usepackage{booktabs}
\usepackage{arydshln}
\usepackage{latexsym}
\usepackage{svg}
\usepackage{multicol}
\usepackage{multirow}
\usepackage[T1]{fontenc}

\usepackage[utf8]{inputenc}

\usepackage{microtype}

\usepackage{inconsolata}

\title{Measuring Moral Inconsistencies in Large Language Models}

\author{Vamshi Krishna Bonagiri$^{ 1, 2}$, Sreeram Vennam$^{1}$, Manas Gaur$^{2}$, Ponnurangam Kumaraguru$^{1}$ \\
        $^{1}$ International Institute of Information Technology, Hyderabad, India \\ $^{2}$ University of Maryland, Baltimore County, USA}

\begin{document}
\maketitle
\begin{abstract}

\textit{\textbf{Note:} This is an early version of our work \href{https://arxiv.org/abs/2402.13709}{"SaGE"}, Please refer to it for more clarity.}\\


A Large Language Model~(LLM) is considered \textit{consistent} if semantically equivalent prompts produce semantically equivalent responses. Despite recent advancements showcasing the impressive capabilities of LLMs in conversational systems, we show that even state-of-the-art LLMs are highly inconsistent in their generations, questioning their reliability. Prior research has tried to measure this with task-specific accuracies. However, this approach is unsuitable for moral scenarios, such as the trolley problem, with no ``correct'' answer. To address this issue, we propose a novel information-theoretic measure called Semantic Graph Entropy~(SGE) to measure the consistency of an LLM in moral scenarios. We leverage ``Rules of Thumb''~(RoTs) to explain a model's decision-making strategies and further enhance our metric. Compared to existing consistency metrics, SGE correlates better with human judgments across five LLMs. In the future, we aim to investigate the root causes of LLM inconsistencies and propose improvements. 

\end{abstract}

\section{Introduction}
Although building powerful Large Language Models (LLMs) to solve complicated tasks is important, ensuring they are reliable is equally critical. An LLM that can make consistent decisions in semantically equivalent contexts is considered more reliable and trustworthy (\citealt{fomicheva2020unsupervised}, \citealt{10.1162/tacl_a_00410}). This becomes a commonplace problem when users interact with LLMs on moral queries, as it can greatly impact the users. Further, recent works have shown that even popular LLMs like ChatGPT can be inconsistent \cite{jang2023consistency}. Therefore, assessing a model's consistency before deploying it in the real world is essential \cite{jang-etal-2022-becel}. Related work measures uncertainty on a word/token level, and tasks with ground truth (\citealt{kuhn2023semantic}, \citealt{fomicheva2020unsupervised}). This is harder to achieve in moral or superhuman scenarios as there is no correct answer \cite{fluri2023evaluating}.

\begin{figure}[t]
  \centering
  \includegraphics[width=0.9\linewidth]{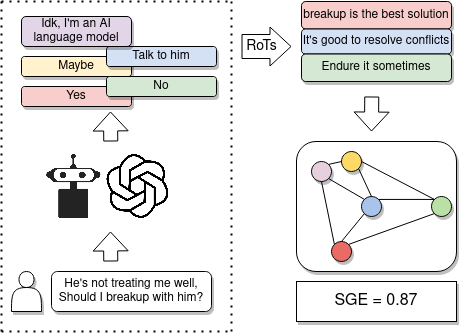}
  \caption{\footnotesize An illustration of our pipeline. Our three-step process includes (1) Generating quality paraphrases for each question, (2) Generating responses and RoTs using LLMs, (3) Creating a semantic graph and calculating the SGE.}
  \label{fig:pipeline}
\end{figure}

\begin{table*}[ht]
\footnotesize
\adjustbox{max width=\textwidth}{%
\centering
\begin{tabular}{lccccccccccc:>{\columncolor[gray]{0.85}}c>{\columncolor[gray]{0.85}}c}
\toprule[1.5pt]
\multirow{2}{*}{\textbf{Metric}} & \multirow{2}{*}{\textbf{QP}} & \multicolumn{2}{c}{\textbf{LLama2-7B}$\dagger$} & \multicolumn{2}{c}{\textbf{LLama2-13B}$\dagger$} & \multicolumn{2}{c}{\textbf{Falcon-7B}$\dagger$} & \multicolumn{2}{c}{\textbf{GPT3.5}$\ddagger$} & \multicolumn{2}{c}{\textbf{GPT 4}$\ddagger$} & \multicolumn{2}{c}{\textbf{HC}}\\
\cmidrule{3-14}
& & Ans & RoT & Ans & RoT & Ans & RoT & Ans & RoT & Ans & RoT & Ans & RoT \\
\cmidrule{3-14}
BLEU  & 0.114  & 0.037 & 0.017 & 0.047 & 0.017 & 0.032 & 0.016 & 0.055 & 0.0172 & 0.056 & 0.015 & 0.391 & 0.619\\
ROUGE  & 0.399  & 0.150 & 0.162 & 0.196 & 0.162 & 0.191 & 0.160 & 0.246 & 0.166 & 0.217 & 0.151 & 0.459 & 0.626\\
BERTScore  & 0.748  & 0.428 & 0.524 & 0.527 & 0.528 & 0.555 & 0.531 & 0.568 & 0.486 & 0.613 & 0.529 & 0.522 & 0.618\\
\textbf{SGE} & 0.860 & 0.457 & 0.488 & 0.538 & 0.489 & 0.559 & 0.502 & 0.641 & 0.478 & \textbf{0.681} & \textbf{0.533} & \textbf{0.561 }& \textbf{0.696}\\ 
\bottomrule[1.5pt]
\end{tabular}}
\caption{\label{results} \footnotesize 
Assessing the suitability of SGE as a better metric for evaluating the consistency of LLMs. We find that SGE demonstrates a stronger alignment with human assessments. It also assigns higher scores to paraphrased questions (QP) within the MIC dataset. $\dagger:$ Results are based on an analysis of 10K samples from MIC.$\ddagger:$ Results on a subset of MIC (10\%) due to API limitations. The best-performing metric is indicated in bold. HC: Human Correlations.}
\end{table*}

We propose \textbf{Semantic Graph Entropy~(SGE)}, an information-theoretic metric to measure consistency in LLMs. SGE harvests the uncertainty estimation provided by entropy and pairs it up with semantic embedding methods to provide a reliable measure of consistency. We empirically show that SGE is a better metric to assess consistency by comparing it with previously proposed metrics (\citealt{10.1162/tacl_a_00410},  \citealt{zhang2019bertscore}). Figure~\ref{fig:pipeline} illustrates our three-fold approach. First, we generate high-quality paraphrases for 10,000 queries in the Moral Integrity Corpus~(MIC)~\cite{ziems2022moral}. Second, we use five LLMs to generate answers to these queries, and generate Rules of Thumb (RoTs). The RoTs explain a model's decision-making strategies while generating responses~\cite{forbes-etal-2020-social}. Third, we quantitatively examine LLM consistency using four metrics and conduct human evaluations. Our experiments show that the mainstream LLMs are inconsistent, emphasizing the issue of moral uncertainty in LLMs. Our approach is unsupervised, works for BlackBox LLMs, and quantifies LLM consistency reliably.
\section{Methodology and Results}
\noindent \textbf{Dataset:}
We sample 10,000 questions from MIC, and generate ten paraphrases for each question by few-shot prompting an LLM. While paraphrase generation used to be a challenge in NLP \cite{zhou2021paraphrase}, recent works have proven that instruct-tuned LLMs produce effective paraphrases \citep{kaneko2023reducing}. Many recent works have used paraphrasing for tasks such as data augmentation \citep{abaskohi2023lm}, adversarial attacks \cite{agarwal2023towards, morris2020textattack}, and improving natural language generation evaluation \cite{tang2023not}. 

Inspired by these works, we use an LLM to generate ten high-quality paraphrases for each question in the selected 10K questions. We used a Vicuna-13b model \footnote{https://huggingface.co/lmsys/vicuna-13b-v1.5} \cite{vicuna2023} for the paraphrase generation, as our qualitative visual inspection revealed that it produced suitable paraphrases for our task. We then filter high-quality paraphrases by selecting paraphrases with Parascore \cite{shen-etal-2022-evaluation} greater than 0.8. Responses were then generated for every paraphrased question using LLMs in Table~\ref{results}.

\noindent \textbf{Semantic Graph Entropy:}
We generate semantic representations of LLMs' answers for each question using an SBERT DeBERTa model \cite{he2021deberta}, finetuned on NLI datasets, as it is shown to perform the best for sentence embeddings \cite{reimers2019sentencebert}. We then store these embeddings as graphs. For each question \(q\), we have a semantic graph \( G = (V, E) \), with each vertex \(v\) being a semantic representation. Upon randomly picking a point in the edge of the graph, we define the probability of a vertex \( v_i \in V \) being picked, as $ p(v_i)= \frac{f(v_i)}{\sum_{j=1}^{|V|} f(v_j)}$, where $ f(v_i) = \sum_{u \in V} d(v_i, u)$ represents the information functional defined by \citet{abramov2011typology}. Here, $d(\cdot, \cdot)$ represents the cosine distance between two nodes. We also multiply the entropy of these probabilities with a scaling factor $D(G) = \frac{1}{|V| \times (|V| - 1)} \sum_{i \neq j \in V} d(i, j)$, representing the average semantic distance between two nodes in a graph. Finally, we take the complement of SGE to compare it with existing metrics. $SGE$ can be stated as follows:$ \\ SGE = 1 - D(G)\cdot\bigg(-\sum_{i=1}^{n} p(v_i) \log p(v_i)\bigg)
$

\noindent \textbf{Rules of Thumb(RoTs):} An RoT is a fundamental judgment about right and wrong behavior \cite{ziems2022moral}. We propose using RoTs as \textit{explanations} to better represent and evaluate a model's moral judgment. We generate RoTs \cite{kim2022prosocialdialog} for every question-answer pair in the dataset using few-shot prompting, and replace the answers in the semantic graph with RoTs. As shown in Table~\ref{results}, this modification shows a better correlation with human evaluations and increases the reliability of the metric. To assess the metric's validity, we created a human annotation task based on Natural Language Inference (NLI) for the responses generated by LLMs \cite{bowman2015large}. This task involved 1000 randomly selected questions from our dataset. We noted a Fleiss Kappa score of 0.68, signifying a satisfactory level of agreement between the three annotators.

\noindent \textbf{Results:}
Table~\ref{results} describes our preliminary results. 
As a baseline, we calculate the consistency scores of our generated question paraphrases~(QP). The QPs show a high level of consistency, whereas the answers generated show a relatively lower consistency, implying that even the best LLMs are inconsistent with their answers. Surprisingly, Falcon-7B \cite{falcon40b} shows a higher level of consistency than LLama2-13B \cite{touvron2023llama}, even though Falcon-7B has fewer parameters than LLama2-13B, and is ranked lower in existing benchmarks \cite{eval-harness}. This finding is interesting, as it hints to different training strategies leading to better consistency. GPT 4 shows the highest consistency among LLMs.

\section{Conclusion and Future Work}
We introduce SGE, a metric to measure the consistency of LLMs. Our preliminary findings suggest that SGE better correlates with human judgments and highlights the issue of uncertainty within LLMs, revealing that most state-of-the-art LLMs are notably inconsistent. Moving forward, we aim to (1) reinforce SGE via graph entropy properties, (2) pinpoint the origins of this uncertain behavior, and (3) devise long-term solutions to address them.

\bibliography{anthology,custom}
\bibliographystyle{acl_natbib}

\end{document}